\documentclass[10pt,conference]{IEEEtran}
\IEEEoverridecommandlockouts
\usepackage{cite}
\usepackage{amsmath,amssymb,amsfonts}
\usepackage{algorithmic}
\usepackage{graphicx}
\usepackage{textcomp}
\usepackage{xcolor}
\usepackage[utf8]{inputenc}
\usepackage{booktabs}
\usepackage{multirow}
\usepackage{hyperref}
\hypersetup{
    colorlinks=true,
    linkcolor=blue,
    citecolor=blue,
    urlcolor=blue
}

\IEEEoverridecommandlockouts
\def\BibTeX{{\rm B\kern-.05em{\sc i\kern-.025em b}\kern-.08em
    T\kern-.1667em\lower.7ex\hbox{E}\kern-.125emX}}
\begin{document}

\title{PropLLM: Propagation-Aware Scene Reconstruction for Network Fault Diagnosis\\
\thanks{This work was supported in part by the National Natural Science Foundation
of China under Grant 62302527, and in part by the High
Performance Computing Center of Central South University. (Corresponding author: Fengxiao Tang.)}
}


\author{
    \IEEEauthorblockN{
        1\textsuperscript{st} Zongzong Wu$^{1}$, 
        2\textsuperscript{nd} Ming Zhao$^{1}$, 
        3\textsuperscript{rd} Fengxiao Tang$^{1}$, 
        4\textsuperscript{th} Nei Kato$^{2}$}
    \IEEEauthorblockA{
        \textsuperscript{1}School of Computer Science and Engineering, \\
        Central South University, Changsha, China \\
        \{Wzy\_Yeah, meanzhao, tangfengxiao\}@csu.edu.cn \\
        \textsuperscript{2}Graduate School of Information Sciences (GSIS), \\
        Tohoku University, Sendai, Japan \\
        nei.kato.d3@tohoku.ac.jp}
}

\maketitle

\begin{abstract}
Network faults propagate layer by layer along topology and protocol dependencies, yet operations systems typically observe only symptomatic alerts at the tail end of propagation chains, where distinct root-cause faults may produce highly similar end-point symptoms. 
  Existing approaches, whether rule-based, machine learning (ML)-based, or large language model (LLM)-based, fundamentally map the alert set to a diagnosis in a single pass and are structurally incapable of resolving this end-point ambiguity. 
This paper proposes PropLLM, which is the first to integrate the hop-by-hop scene reconstruction paradigm with the generative reasoning capabilities of LLMs. 
   Starting from end-point alerts, PropLLM traces back hop-by-hop along the propagation path, retrieving verifiable factual evidence from a dual-layer knowledge graph (KG) at each hop, while the proposed Temporal Causal Propagation Attention (TCPA) mechanism encodes known topological causal priors directly into the attention computation to guide the model along the correct causal direction, ultimately localizing the root cause and determining the fault type through a fully evidenced causal chain. 
On a real-world Wi-Fi multimodal fault dataset, PropLLM improves fault type diagnosis accuracy by 3.9\% and root cause localization accuracy by 4.7\% over the strongest baseline, while reducing the hallucination rate by 50.8\%. Supplementary experiments on the TeleLogs 5G dataset further demonstrate the effectiveness of the proposed method across different network scenarios.
\end{abstract}


\section{Introduction}
\begin{figure}[t!]
\centering
\includegraphics[width=\columnwidth]{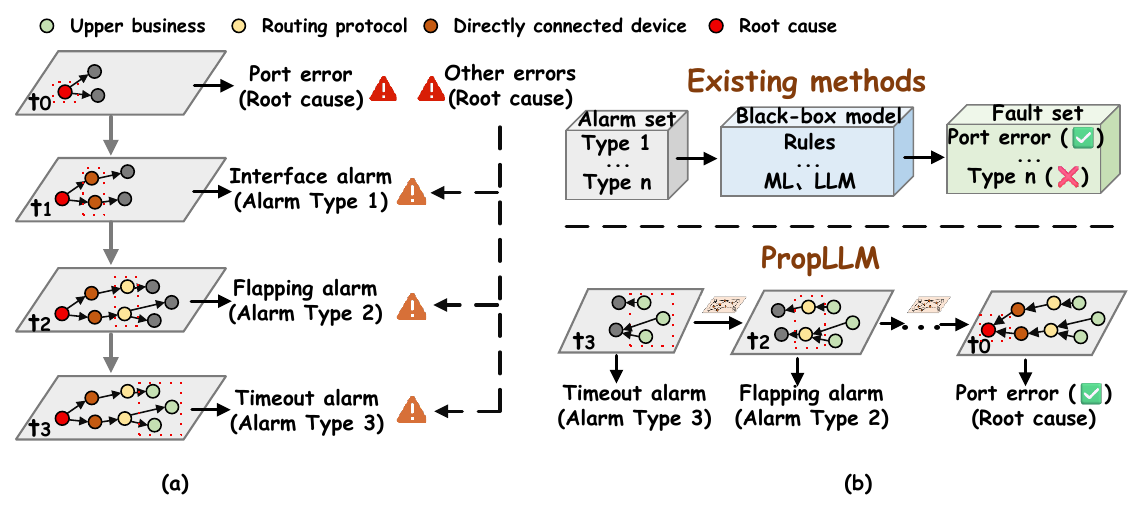}
\caption{(a) Fault cascading propagation produces ambiguous end-point alerts. (b) Top: one-shot mapping paradigm; Bottom: hop-by-hop scene reconstruction paradigm of PropLLM.}
\label{F1-INTRO.pdf}
\end{figure}

Network fault diagnosis (NFD) aims to accurately determine fault types (e.g., link degradation, misconfiguration, device failure) from observed anomalies, thereby guiding operators toward targeted remediation~\cite{steinder2004survey,zhang2024survey}. The fundamental difficulty of this task lies in the fact that network faults propagate layer by layer along topology and protocol dependencies, yet operations systems typically observe only symptomatic alerts at the tail end of propagation chains (as illustrated in Fig.~\ref{F1-INTRO.pdf}(a)). These end-point alerts are highly ambiguous: different types of root-cause faults may produce similar downstream symptoms, and the same fault may manifest entirely different alert patterns along different propagation paths.

Current mainstream methods—whether based on rule matching~\cite{steinder2004survey}, machine learning classification~\cite{zhao2019deep,deng2021gdn}, or large language model (LLM)-based generation~\cite{ahmed2023recommending}—all follow a single-pass mapping paradigm. They directly map the observed alert set to a fault type or root cause through one-step feature extraction or reasoning, without tracing backward along the propagation path for hop-by-hop reconstruction and hypothesis verification (as shown in the upper half of Fig.~\ref{F1-INTRO.pdf}(b)).
The core limitation of this paradigm is that alerts represent only the end products of fault propagation. The same set of symptoms can arise from heterogeneous root causes via different paths, making it impossible to resolve endpoint ambiguity from observations alone. Accurate fault diagnosis requires reconstructing the full causal chain by tracing back to the root cause, which the single-pass paradigm is inherently unable to achieve.

In practice, experienced network engineers resolve end-point alert ambiguity by tracing back hop-by-hop along topology and protocol dependencies. They reconstruct node states, validate causal hypotheses, and localize the root cause (as shown in the lower half of Fig.~\ref{F1-INTRO.pdf}(b)). This approach ensures diagnosis is based on a fully evidenced causal chain rather than statistical symptom correlations.
Automating this process requires a reasoning engine that interprets multimodal data, dynamically adjusts hypotheses, and produces interpretable diagnosis chains~\cite{trivedi2023interleaving}. LLMs are a natural choice for this task due to their cross-modal understanding and generative reasoning capabilities~\cite{wang2025llm,wang2024netassistant,ahmed2023recommending,chen2024rca}.
Integrating LLMs into hop-by-hop reconstruction requires two conditions: (1) verifiable evidence must be retrievable at each hop for state reconstruction, and (2) the reasoning must be causally aware, distinguishing upstream causes from downstream effects.

However, existing work has systematic gaps in factual grounding, making hop-by-hop scene reconstruction difficult in practice. Scene reconstruction requires restoring each node’s state using two distinct types of facts: structural knowledge (topology, protocol configurations, and device parameters) and experiential knowledge (historical fault patterns on similar paths).
Although knowledge graphs (KGs) are suitable for organizing both~\cite{ji2022survey}, most existing methods use flat structures that mix them indiscriminately~\cite{liu2022concurrent,wang2024comprehensive}. This mixing introduces retrieval noise and fails to capture hierarchical relationships, as the two knowledge types differ in update frequency, query patterns, and indexing~\cite{hu2020heterogeneous}. Moreover, current RAG methods~\cite{lewis2020retrieval} inject knowledge in a single pass~\cite{asai2024selfrag}, lacking hop-by-hop verification, which makes models prone to hallucination and produces diagnosis paths that deviate from the true causal chain.

Existing work also lacks causal awareness. Standard attention mechanisms~\cite{vaswani2017attention} assign symmetric weights to all tokens, failing to distinguish causal direction and often misidentifying end-point symptoms as root causes. Although chain-of-thought (CoT) prompting~\cite{wei2022chain} can guide reasoning steps, it does not alter the underlying symmetric attention computation, so directional errors persist in complex scenarios.
Crucially, network fault propagation direction is a known structural prior derived from topology and protocol dependencies, unlike microservice systems where causality must be discovered from data~\cite{ikram2022root,yu2023nezha,yao2024chain}. The key missing component is a mechanism that encodes these causal priors directly into attention layers, enabling directional awareness at every step rather than relying on post-hoc prompts.

This paper proposes PropLLM, the first framework to integrate hop-by-hop hypothesis-verification scene reconstruction with the generative reasoning capabilities of LLMs. To enable factual grounding, we construct a dual-layer knowledge graph that separates structural knowledge of topology and protocols from experiential knowledge of historical fault patterns, with cross-layer associations for efficient retrieval. To achieve propagation-aware causal reasoning, we introduce the Temporal Causal Propagation Attention (TCPA) mechanism, which encodes topological causal priors into every attention layer through a causal direction mask, propagation diffusion matrix, and temporal bias. The TCPA output is injected into the LLM decoder via cross-attention. Built on this foundation, PropLLM performs hop-by-hop reconstruction through a dynamic closed loop, where verification results trigger targeted retrieval at the next hop, constraining the reasoning chain with factual evidence and effectively suppressing hallucination.

The main contributions of this paper are as follows:

\begin{itemize}
\item  We propose the PropLLM framework, which for the first time formalizes the hop-by-hop scene reconstruction methodology of human experts into a computable reasoning paradigm, revealing that accurate fault type diagnosis depends on a fully evidenced causal chain rather than single-pass mapping from end-point observations.

\item  We propose TCPA, a standalone Transformer encoder that encodes known topological causal priors into every attention layer and injects its output into the LLM Decoder via cross-attention, enabling continuous perception of fault propagation direction during generation.

\item We construct a dual-layer KG separating structural from experiential knowledge, with a dynamic closed-loop retrieval mechanism that constrains each hop's reasoning with factual evidence, effectively suppressing hallucination.
\end{itemize}

\section{RELATED WORK}
\subsection{Network Fault Diagnosis Methods} 
The evolution of NFD methods has shifted from rule-driven to data-driven approaches. Rule-based methods~\cite{steinder2004survey} depend on expert-defined rules but scale poorly and cover only known faults. Traditional ML methods~\cite{steinder2004survey,zhao2019deep} treat diagnosis as classification but struggle with topological and temporal dependencies. Deep learning approaches~\cite{deng2021gdn,li2024exchain,zhang2024illuminating} improve representation learning yet still rely on statistical co-occurrence and falter under ambiguous alert patterns from multiple root causes.
Unlike microservice systems (e.g., MULAN~\cite{zheng2024mulan}, Minder~\cite{deng2025minder}), where causality must be discovered from data, network fault propagation is a known structural prior from topology and protocols. Recent LLM-based methods~\cite{ahmed2023recommending,chen2024rca,jiang2024xpert,wang2025llm} show promise, but effectively harnessing their reasoning capabilities in knowledge interaction and causal awareness remains an open question.

 \subsection{LLM-Driven Fault Diagnosis}
Existing work on applying LLMs to fault diagnosis can be categorized into three main approaches.
The end-to-end generation approach directly maps event descriptions to diagnostic results~\cite{ahmed2023recommending,chen2024rca,zhang2024lmpace}. While offering strong zero-shot generalization, these methods rely solely on parametric knowledge and lack external factual grounding.
The retrieval-augmented approach enhances LLMs via RAG~\cite{lewis2020retrieval}~\cite{jiang2024xpert,wang2025llm}. However, they typically follow a single-retrieval, single-generation pipeline, which cannot support the dynamic knowledge demands of multi-hop reasoning.
The agent-based approach equips LLMs with dynamic tool invocation~\cite{wang2024netassistant,yao2024chain}. Nevertheless, they still suffer from temporal decoupling between knowledge acquisition and reasoning, preventing closed-loop hop-by-hop verification.
Furthermore, all three approaches lack causal awareness. LLM self-attention~\cite{vaswani2017attention} assigns symmetric weights and cannot distinguish causal direction, while CoT prompting~\cite{wei2022chain} only operates at the decoding level.
In summary, existing LLM-based fault diagnosis methods suffer from two core limitations: (1) insufficient on-demand hop-by-hop knowledge interaction and verification, and (2) inability to perceive causal propagation direction in attention mechanisms.

\subsection{Knowledge Graph Representation and Causal Reasoning}
The previous subsection identified knowledge interaction and causal awareness as two core gaps. This subsection reviews the technical foundations of the proposed dual-layer KG and TCPA.

For KG representation in fault diagnosis, early studies mainly relied on single-layer knowledge graphs combined with GCNs for fault classification~\cite{liu2022concurrent}. Recent methods introduced temporal knowledge graphs in UniDiag~\cite{zhang2024unidiag} and hierarchical structures in KG4Diagnosis~\cite{zuo2025kg4diagnosis}. KG-augmented LLM approaches~\cite{wu2026kgv,luo2024rog,fang2025kirag} further explore retrieval-reasoning closed loops. However, none of these methods separates structural knowledge from experiential knowledge into semantically linked yet distinct layers. This separation is critical for accurate fault diagnosis, as it enables simultaneous verification of current states using structural knowledge and historical fault patterns using experiential knowledge at each reasoning step. Mixing the two types introduces retrieval noise and hinders precise on-demand verification.

For causally-aware attention, existing methods~\cite{kong2024causalformer,vashishtha2025causal,cheng2024cuts,rohekar2023causal} focus on discovering latent causal relations from data. In contrast, PropLLM operates under a known structural prior of fault propagation direction obtainable from network configurations. While input-stage GNNs inject topology only once, TCPA encodes causal priors into every attention layer through its causal direction mask and propagation diffusion matrix, achieving deeper integration.

\section{Dual-Layer Knowledge Graph Construction}
\begin{figure}[t]
  \centering
  \includegraphics[width=\columnwidth]{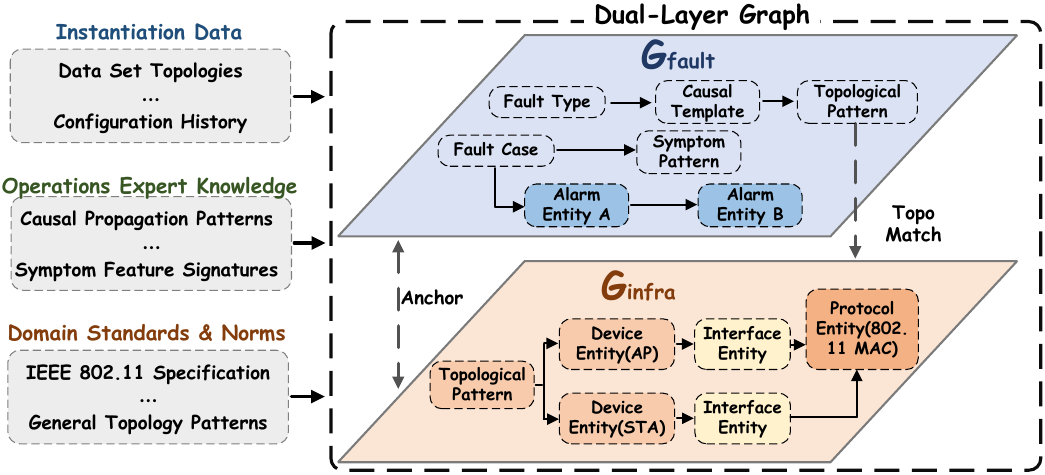}
  \caption{Architecture of the dual-layer knowledge graph with cross-layer semantic links.}
  \label{F2.pdf}
\end{figure}

Each step of hop-by-hop verification requires two fundamentally different types of knowledge, namely structural knowledge describing network topology and protocol behavior specifications, and experiential knowledge derived from historical fault cases. We organize these into an infrastructure-layer graph $\mathcal{G}_{\text{infra}}$ and a fault-experience-layer graph $\mathcal{G}_{\text{fault}}$, linked by cross-layer semantic associations for joint querying. Fig.~\ref{F2.pdf} illustrates the overall architecture and knowledge sources of the dual-layer KG.

The knowledge sources of the dual-layer KG fall into three categories: (1)~domain standards (IEEE~802.11, TCP/IP specifications, common topology patterns) from protocol documents and engineering textbooks, forming the backbone of $\mathcal{G}_{\text{infra}}$; (2)~operations expert knowledge (causal propagation patterns, symptom signatures, diagnostic rules) from expert interviews and manuals, constituting prior causal templates in $\mathcal{G}_{\text{fault}}$; and (3)~instantiation data (topology configurations, monitoring records) from specific network environments, binding generic knowledge to a concrete network.

Since the third category partially overlaps with the evaluation dataset, we clarify fairness safeguards. Information from the dataset in $\mathcal{G}_{\text{infra}}$ is limited to topology structure and static device attributes---equivalent to a deployment-time asset inventory without fault labels or runtime state, thus not constituting information leakage. Case knowledge in $\mathcal{G}_{\text{fault}}$ is drawn strictly from the training split, with the first 80\% by occurrence time used for graph construction and no test case participating. Protocol specifications and topology patterns in $\mathcal{G}_{\text{infra}}$, as well as prior causal templates in $\mathcal{G}_{\text{fault}}$, are publicly available domain knowledge independent of any specific dataset.

\subsection{Infrastructure-Layer Graph $\mathcal{G}_{\text{infra}}$}

$\mathcal{G}_{\text{infra}}$ structurally encodes the network's topology, device capabilities, protocol behavior specifications, and communication patterns, providing hop-by-hop verification with foundational facts such as each device's role, connectivity, protocol configuration, and normal behavior baseline.

Let $G_\text{infra} = (\mathcal{V}_\text{infra}, \mathcal{E}_\text{infra}, \mathcal{A}_\text{infra})$, where $\mathcal{V}_\text{infra}$ is the entity set, $\mathcal{E}_\text{infra} \subseteq \mathcal{V}_\text{infra} \times \mathcal{R}_\text{infra} \times \mathcal{V}_\text{infra}$ is the relation set, $\mathcal{R}_\text{infra}$ is the relation type set, and $\mathcal{A}_\text{infra}: \mathcal{V}_\text{infra} \to 2^{\mathcal{K} \times \text{Val}}$ is the attribute mapping.

$\mathcal{V}_\text{infra}$ comprises four entity types: \emph{device} entities carrying role labels that determine structural positions along propagation paths; \emph{interface} entities recording operating mode, frequency band, and channel parameters; \emph{protocol} entities encoding behavioral specifications serving as anomaly detection baselines; and \emph{topology pattern} entities representing common network topologies and their propagation characteristics. $\mathcal{R}_\text{infra}$ contains six relation types: physical/wireless connectivity, device--interface ownership, protocol instantiation, data-flow communication, topology pattern instantiation, and inter-protocol dependency.

$G_\text{infra}$ is constructed in two stages: \emph{generic knowledge injection} extracts protocol entities, topology patterns, and inter-protocol dependencies from standard documents and engineering knowledge bases; \emph{instantiation} binds this generic knowledge to a concrete network by extracting devices and connectivity from topology configurations and parsing interface attributes and communication relations from device and traffic records.

\subsection{Fault-Experience-Layer Graph $\mathcal{G}_{\text{fault}}$}

$\mathcal{G}_{\text{fault}}$ structurally encodes the causal propagation patterns of fault types and the observational data of historical cases, providing hop-by-hop verification with experiential evidence on how specific symptoms typically propagate and what root causes they have historically pointed to.

Let $G_\text{fault} = (\mathcal{V}_\text{fault}, \mathcal{E}_\text{fault}, \mathcal{A}_\text{fault})$, where $\mathcal{V}_\text{fault}$ is the fault-related entity set, $\mathcal{E}_\text{fault} \subseteq \mathcal{V}_\text{fault} \times \mathcal{R}_\text{fault} \times \mathcal{V}_\text{fault}$ is the relation set, and $\mathcal{A}_\text{fault}$ is the attribute mapping.

$\mathcal{V}_\text{fault}$ comprises five entity types: \emph{fault type} entities encoding definitions and triggering conditions; \emph{causal template} entities describing typical propagation paths distilled from expert knowledge; \emph{fault case} entities carrying type labels, timestamps, and root-cause devices; \emph{alert} entities with codes, severity levels, and device identifiers; and \emph{symptom pattern} entities aggregating similar alert combinations across cases. $\mathcal{R}_\text{fault}$ contains six relation types: fault type--template association, temporal triggering between alerts, alert--case membership, case--type determination, case--device localization, and case--symptom pattern association.

The temporal triggering relation is the core structure of $G_\text{fault}$. For each training-set case, alerts are sorted by timestamp; for pairs within a time window $\Delta t_\text{max}$ whose devices are topologically connected in $G_\text{infra}$, a triggering edge is established from earlier to later, and transitive reduction removes redundant edges, converting each case into a directed acyclic propagation graph. Symptom pattern entities are generated by hierarchical clustering over training-set cases using alert code sets as features, each associated with a fault type distribution for rapid hypothesis narrowing.

\subsection{Cross-Layer Semantic Links}

The core value of the dual-layer KG lies in cross-layer joint querying. When traceback reaches a device, the model must simultaneously access its infrastructure attributes and historical propagation behavior under similar alerts. This is realized through two semantic links: \emph{device anchoring} links alert entities in $\mathcal{G}_{\text{fault}}$ to device entities in $\mathcal{G}_{\text{infra}}$ via identifiers, enabling retrieval of both historical alerts and topological position in a single traversal; \emph{topology-template association} connects causal templates in $\mathcal{G}_{\text{fault}}$ with topology patterns in $\mathcal{G}_{\text{infra}}$, allowing the model to filter inapplicable templates based on the current network's topology.

The complete dual-layer KG is defined as $\mathcal{G} = (\mathcal{G}_{\text{infra}}, \mathcal{G}_{\text{fault}}, \mathcal{L}_{\text{cross}})$, where $\mathcal{L}_{\text{cross}}$ is the cross-layer link set. For the $k$-th step of hop-by-hop verification, given the current device $d_k$ and causal hypothesis $h_k$, the model executes the following cross-layer joint queries:

\begin{align}
  q_{\text{struct}}(d_k) &= \{(v, r, v') \in \mathcal{E}_{\text{infra}}
  \mid v = d_k \lor v' = d_k\}
  \end{align}
  \begin{multline}
  q_{\text{exp}}(d_k, h_k) = \{(v, r, v') \in \mathcal{E}_{\text{fault}}
  \mid \text{anchor}(v) = d_k \\
  \land\; \text{sim}(\text{attr}(v), h_k) > \tau\}
\end{multline}
where $q_{\text{struct}}$ returns all associated facts of device $d_k$ in the infrastructure layer, including topological neighbors, protocol configurations, and normal communication baselines; $q_{\text{exp}}$ returns historical alerts and propagation records of device $d_k$ in the fault-experience layer that are semantically similar to the current hypothesis $h_k$, with $\tau$ as the similarity threshold. The results of both queries jointly constitute the factual evidence for verification at step $k$.

\subsection{Graph Statistics and Analysis}

Table~\ref{tab:kg_stats} reports the KG statistics. Protocol and topology pattern entities in $\mathcal{G}_{\text{infra}}$ originate from public domain knowledge (55.8\% of $\mathcal{G}_{\text{infra}}$ entities), ensuring transferability---deploying to a new network requires updating only device and interface entities. $\mathcal{G}_{\text{fault}}$ accounts for 95.5\% of total entities and 77.6\% of relations, reflecting that diagnostic complexity lies in propagation pattern diversity rather than network structure. Every alert entity is anchored to a device entity for seamless cross-layer access. All cases in $\mathcal{G}_{\text{fault}}$ are strictly confined to the training split.

 \begin{table}[h]
  \centering
  \caption{Dual-layer knowledge graph statistics.}
  \label{tab:kg_stats}
  \resizebox{\columnwidth}{!}{
  \begin{tabular}{llrl}
  \toprule
  \textbf{Layer} & \textbf{Entity Type} & \textbf{Count} & \textbf{Source} \\
  \midrule
  \multirow{5}{*}{$\mathcal{G}_{\text{infra}}$}
   & Device       & 21     & Dataset topology config \\
   & Interface    & 21     & Dataset device records \\
   & Protocol     & 47     & Protocol standards (IEEE/IETF) \\
   & Topology pattern & 6  & Network engineering KB \\
   & Relations    & 312    & --- \\
  \midrule
  \multirow{6}{*}{$\mathcal{G}_{\text{fault}}$}
   & Fault type   & 11     & Domain taxonomy \\
   & Causal template & 34  & Operations expert knowledge \\
   & Fault case   & 472    & Training split (80\%) \\
   & Alert        & 14,780 & Training split (80\%) \\
   & Symptom pattern & 89  & Clustering over training cases \\
   & Relations    & 52,436 & --- \\
  \midrule
  \multirow{2}{*}{$\mathcal{L}_{\text{cross}}$}
   & Anchor       & 14,780 & Device identifier matching \\
   & Topo\_match  & 34     & Topology-template semantic matching \\
  \midrule
  \multicolumn{2}{l}{\textbf{Total entities}} & 15,481 & \\
  \multicolumn{2}{l}{\textbf{Total relations}} & 67,562 & \\
  \bottomrule
  \end{tabular}}
\end{table}

\section{PropLLM Framework}

\begin{figure*}[t]
  \centering
  \includegraphics[width=\textwidth]{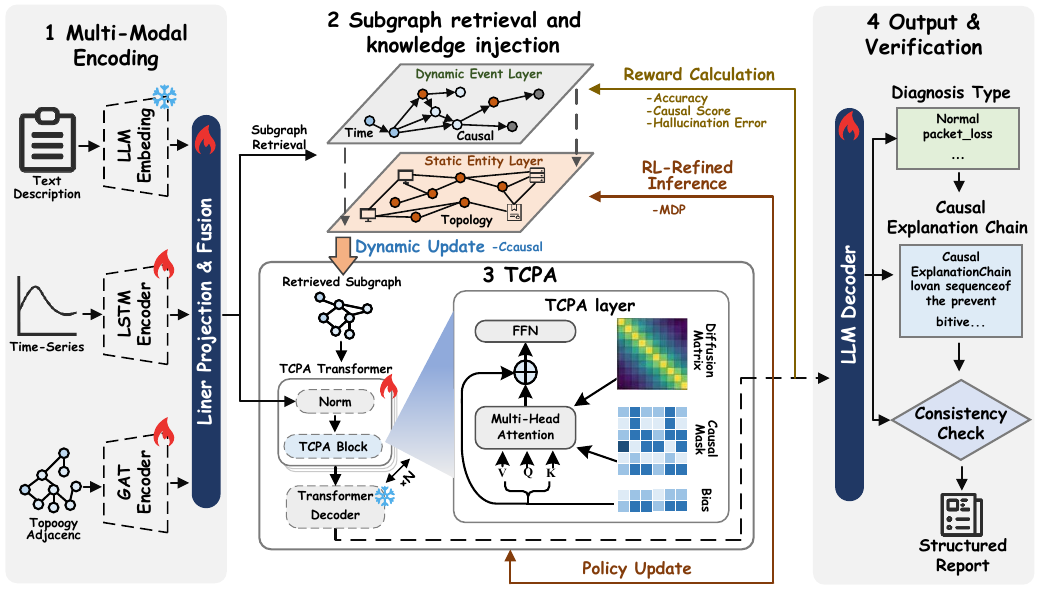}
  \caption{Overall architecture of the PropLLM framework.}
  \label{F3.pdf}
  \end{figure*}

As illustrated in Fig.~\ref{F3.pdf}, PropLLM consists of four modules: (1)~multimodal encoding, which maps heterogeneous monitoring data into a unified semantic space; (2)~subgraph retrieval and knowledge injection, which dynamically retrieves relevant subgraphs from the dual-layer KG based on the current reasoning state; (3)~the TCPA Transformer, which drives hop-by-hop causal traceback along the propagation path; and (4)~output and verification, which generates fault type diagnoses and causal explanation chains with consistency checking. Training employs reinforcement learning with diagnosis accuracy, causal chain quality, and hallucination suppression as joint reward signals.

\subsection{Multimodal Encoding Module}

The input comprises three modalities: alert text/logs, device performance time series, and topological graph structure. Three parallel encoding paths capture modality-specific structures.

For alert and log text of device $d_i$, a pretrained language model embedding layer~\cite{devlin2019bert} produces:
\begin{equation}
  \mathbf{h}_i^\text{text} = \text{LLM\_Embed}(\text{Concat}(w_1, \ldots, w_m)) \in \mathbb{R}^{d_t}
\end{equation}
where $\{w_1, \ldots, w_m\}$ is the concatenated alert sequence and $d_t$ is the embedding dimension.

For performance time series $\mathbf{X}_i \in \mathbb{R}^{T \times F}$ with $T$ time steps and $F$ features, a bidirectional LSTM~\cite{hochreiter1997long} captures temporal dependencies:
\begin{equation}
  \mathbf{h}_i^\text{ts} = \text{LSTM}(\mathbf{X}_i; \theta_\text{lstm}) \in \mathbb{R}^{d_s}
\end{equation}
where $\mathbf{h}_i^\text{ts}$ is the final hidden state and $d_s$ is the hidden dimension.

For the network topology with adjacency matrix $\mathbf{A} \in \mathbb{R}^{N \times N}$, a graph attention network (GAT)~\cite{velickovic2018graph} encodes topological context:
\begin{equation}
  \mathbf{h}_i^\text{topo} = \text{GAT}(\mathbf{H}^{(0)}, \mathbf{A}; \theta_\text{gat}) \in \mathbb{R}^{d_g}
\end{equation}
where $\mathbf{H}^{(0)}$ is the initial feature matrix from device role encodings.

The three representations are fused via learnable linear projections:
\begin{equation}
  \mathbf{z}_i = \mathbf{W}_t \mathbf{h}_i^\text{text} + \mathbf{W}_s \mathbf{h}_i^\text{ts} + \mathbf{W}_g
  \mathbf{h}_i^\text{topo} + \mathbf{b}
\end{equation}
where $\mathbf{W}_t \in \mathbb{R}^{d \times d_t}$, $\mathbf{W}_s \in \mathbb{R}^{d \times d_s}$, $\mathbf{W}_g \in \mathbb{R}^{d \times d_g}$ are projection matrices. The fused matrix $\mathbf{Z} = [\mathbf{z}_1, \ldots, \mathbf{z}_N]^\top \in \mathbb{R}^{N \times d}$ serves as input to subsequent stages.

\subsection{Subgraph Retrieval and Knowledge Injection}

Hop-by-hop verification requires the model to retrieve knowledge relevant to the current causal hypothesis at each reasoning step. At the $k$-th step of the TCPA Transformer, the retrieval query is formulated as:

\begin{equation}
  \text{query}_k = (d_k,\; \text{code}_k,\; \mathbf{s}_k)
\end{equation}
where $d_k$ is the device currently reached by traceback, $\text{code}_k$ is its alert code set, and $\mathbf{s}_k$ is the model's current hidden state. These three components respectively specify the spatial anchor, content anchor, and semantic anchor of retrieval.

Based on this query, the model performs cross-layer subgraph retrieval on the dual-layer KG. On $\mathcal{G}_{\text{infra}}$, an $L$-hop neighborhood expansion centered at $d_k$ extracts the structural subgraph:

\begin{equation}
  \mathcal{G}_k^{\text{infra}} = \text{$L$-hop-subgraph}
  (\mathcal{G}_{\text{infra}},\; d_k)
\end{equation}
which contains the device's topological neighbors, interface attributes, protocol configurations, and normal communication baselines. On $\mathcal{G}_{\text{fault}}$, the anchoring link locates historical alert entities of $d_k$, filters the subset matching $\text{code}_k$, and expands along triggering edges to extract the experiential subgraph:

\begin{equation}
  \mathcal{G}_k^{\text{fault}} = \text{PropSubgraph}
  (\mathcal{G}_{\text{fault}},\; \text{anchor}(d_k),\; \text{code}_k)
\end{equation}
which contains related propagation path segments and root-cause information. The two subgraphs are merged as $\mathcal{G}_k = \mathcal{G}_k^{\text{infra}} \cup \mathcal{G}_k^{\text{fault}}$.

The retrieved subgraph is converted into a sequence via structure-aware linearization~\cite{chen2020kgpt}: each triple $(h, r, t)$ in $\mathcal{G}_k$ is rendered as a natural language description with structural markers and encoded through the LLM embedding layer, yielding the knowledge matrix $\mathbf{K}_k \in \mathbb{R}^{|\mathcal{G}_k| \times d}$, injected into the TCPA layer via cross-attention. Unlike single-pass RAG~\cite{lewis2020retrieval}, the query at each step is dynamically determined by the model's reasoning state, so retrieval focus automatically shifts from end-point to upstream devices as traceback progresses, achieving hop-by-hop synchronization between reasoning and knowledge acquisition.

\subsection{Temporal Causal Propagation Attention (TCPA)}

TCPA enables PropLLM to perceive fault propagation direction at the attention level. Standard self-attention~\cite{vaswani2017attention} assigns symmetric weights and cannot distinguish causal upstream from downstream. TCPA addresses this through three mechanisms: a causal direction mask constraining information flow, a propagation diffusion matrix encoding causal reachability, and a temporal propagation bias capturing event temporal order.

\subsubsection{Causal Direction Mask}
Fault propagation spreads from the root-cause device downstream along topology and protocol dependencies, so hop-by-hop traceback should proceed in the reverse direction from end-point alerts toward the upstream. The causal direction mask $\mathbf{M}_{\text{causal}} \in \mathbb{R}^{N \times N}$ enforces this prior by constraining attention directionality:

\begin{equation}
  \mathbf{M}_{\text{causal}}(i,j) =
  \begin{cases}
  0, & \text{if } j \text{ is causal upstream or peer of } i \\
  -\infty, & \text{otherwise}
  \end{cases}
\end{equation}
Causal upstream is determined jointly by topological distance and alert temporal order. Device $j$ is the causal upstream of device $i$ if and only if:

\begin{equation}
  \label{eq:upstream}
  \text{upstream}(j, i) \equiv
  (\text{depth}(j) \le \text{depth}(i)) \land
  (t_j^{\text{first}} \le t_i^{\text{first}})
\end{equation}
where $\text{depth}(\cdot)$ denotes the device's depth in the topology tree (AP as root, closer to root is upstream) and $t_{\cdot}^{\text{first}}$ denotes its earliest alert timestamp. Positions masked with $-\infty$ yield near-zero attention after softmax, blocking reverse information flow and forcing the model to trace back exclusively toward the root cause.

Note that $\text{depth}(\cdot)$ in Eq.~(\ref{eq:upstream}) is a concrete instantiation for tree topologies, not a structural constraint of TCPA. For general topologies, it can be replaced by any partial order function reflecting causal propagation direction---e.g., topological sort on forwarding DAGs in mesh networks, or inter-base-station interference order in the TeleLogs 5G scenario (validated in our experiments). When the topological partial order is locally undefined (e.g., routing loops), the temporal constraint $t_j^\text{first} \leq t_i^\text{first}$ provides a fallback cue. When both degenerate simultaneously, TCPA sets the mask to zero rather than $-\infty$, gracefully degrading to standard self-attention for that device pair.

\subsubsection{Propagation Diffusion Matrix}

The causal direction mask provides binary directional constraints, but the causal association strength at different positions along the propagation path is not uniformly distributed; upstream events closer to the root cause typically carry stronger causal signals. The propagation diffusion matrix $\mathbf{D} \in \mathbb{R}^{N \times N}$ continuously encodes this non-uniform causal strength.

We construct $\mathbf{D}$ based on the graph diffusion process~\cite{gasteiger2019diffusion}. Given the normalized graph Laplacian $\mathbf{L} = \mathbf{I} - \mathbf{D}_{\text{deg}}^{-1/2} \mathbf{A} \mathbf{D}_{\text{deg}}^{-1/2}$, where $\mathbf{D}_{\text{deg}}$ is the degree matrix, the diffusion matrix is computed via the matrix exponential with $K$-th order truncation:

\begin{equation}
  \mathbf{D} = \exp(-\beta \mathbf{L}) \approx
  \sum_{k=0}^{K} \frac{(-\beta \mathbf{L})^k}{k!}
\end{equation}
where $\beta > 0$ controls the propagation decay rate and $K$ is the truncation order. $\mathbf{D}(i,j)$ reflects diffusion reachability from device $j$ to $i$, decaying exponentially with hop count. This matrix encodes three properties: maximum diffusion between directly connected devices, cumulative effects when multiple paths exist, and exponential decay with distance that prioritizes nearest upstream events while suppressing distant noise.

The truncation order $K{=}4$ covers the maximum propagation depth in the dataset. Each attention head uses an independent $\beta_h$ whose initialization is determined by grid search, enabling different heads to attend to propagation scales from local 1--2 hops to longer-range chains.

\subsubsection{Temporal Propagation Bias}

The direction mask and diffusion matrix are both based on static topology and do not exploit alert timestamp information. However, alert pairs at the same topological distance may have vastly different temporal intervals, and temporally proximate pairs are more likely to lie on the same propagation chain. The temporal propagation bias $\mathbf{B} \in \mathbb{R}^{N \times N}$ encodes this temporal signal as attention biases:

\begin{equation}
  \mathbf{B}(i,j) = -\lambda \cdot
  \frac{|t_i^{\text{first}} - t_j^{\text{first}}|}{\Delta t_{\max}}
\end{equation}
where $\lambda > 0$ is the temporal decay coefficient and $\Delta t_{\max}$ is the maximum alert time span serving as normalization. This bias assigns higher weights to temporally proximate alert pairs while attenuating those with large gaps. Unlike positional encodings~\cite{vaswani2017attention}, $\mathbf{B}(i,j)$ is based on actual event times rather than sequence positions, correctly handling the non-uniform temporal distribution of fault propagation.

\subsubsection{TCPA Attention Computation}

The three components are integrated into the attention computation to form the complete TCPA:

\begin{equation}
  \text{TCPA}(\mathbf{Q}, \mathbf{K}, \mathbf{V}) =
  \text{softmax}\!\left(
  \frac{\mathbf{Q}\mathbf{K}^\top}{\sqrt{d_k}} \odot \mathbf{D}
  + \mathbf{B} + \mathbf{M}_{\text{causal}}
  \right) \mathbf{V}
\end{equation}
where $\mathbf{Q} = \mathbf{Z}\mathbf{W}_Q$, $\mathbf{K} = \mathbf{Z}\mathbf{W}_K$, $\mathbf{V} = \mathbf{Z}\mathbf{W}_V$ are query, key, and value projections, $\odot$ denotes element-wise multiplication, and $d_k$ is the scaling factor. Compared to standard self-attention $\text{softmax}(\mathbf{Q}\mathbf{K}^\top / \sqrt{d_k})\mathbf{V}$, TCPA introduces a triple causal inductive bias: $\mathbf{D}$ weights attention scores by propagation diffusion strength, $\mathbf{B}$ adjusts biases by temporal proximity, and $\mathbf{M}_{\text{causal}}$ blocks attention flow that violates the causal direction.

TCPA employs multi-head attention~\cite{vaswani2017attention} with each head $h$ using an independent $\beta_h$, allowing low-$\beta$ heads to focus on tightly coupled local causal pairs while high-$\beta$ heads capture long-range propagation effects across multiple hops. Following multi-head self-attention, cross-attention integrates the knowledge representation $\mathbf{K}_k$ from the subgraph retrieval module:

\begin{equation}
  \text{CrossAttn}(\mathbf{Z}', \mathbf{K}_k) =
  \text{softmax}\!\left(
  \frac{\mathbf{Z}'\mathbf{W}_Q^c (\mathbf{K}_k \mathbf{W}_K^c)^\top}
  {\sqrt{d_k}}
  \right) \mathbf{K}_k \mathbf{W}_V^c
\end{equation}
where $\mathbf{Z}'$ is the output of multi-head self-attention. This cross-attention enables the model to selectively attend to the most relevant facts in the retrieved knowledge for verifying the current causal hypothesis at each reasoning step.

The complete TCPA block follows the Pre-Norm architecture~\cite{xiong2020layer}:

\begin{align}
  \mathbf{Z}' &= \mathbf{Z} + \text{MultiHead}(\text{LN}(\mathbf{Z})) \\
  \mathbf{Z}'' &= \mathbf{Z}' +
  \text{CrossAttn}(\text{LN}(\mathbf{Z}'), \mathbf{K}_k) \\
  \mathbf{Z}_{\text{out}} &= \mathbf{Z}'' +
  \text{FFN}(\text{LN}(\mathbf{Z}''))
\end{align}
where FFN is a two-layer feed-forward network and LN denotes LayerNorm. TCPA blocks are stacked for $L$ layers, sharing $\mathbf{M}_{\text{causal}}$ and $\mathbf{B}$ across layers (as these are statically determined by the input event) while independently learning diffusion coefficients and projection parameters at each layer.

The output $\mathbf{Z}^{(L)}$ of the TCPA Transformer is injected as a prefix sequence into the cross-attention layers of the LLM Decoder, enabling the Decoder to continuously perceive the causal structure encoded by TCPA during autoregressive generation. As reasoning unfolds hop-by-hop, the subgraph retrieval module provides updated knowledge $\mathbf{K}_{k+1}$, and the cross-attention in the TCPA layer switches to the new knowledge context accordingly, forming an iterative closed loop of traceback, retrieval, and verification.

\subsection{Output and Verification Module}
The LLM Decoder generates a diagnostic sequence that is mapped to a fault type probability distribution through a classification head:

\begin{equation}
  p(y | \mathbf{X}) = \text{softmax}(\mathbf{W}_{\text{cls}}
  \mathbf{h}_{\text{[CLS]}} + \mathbf{b}_{\text{cls}})
\end{equation}
where $\mathbf{h}_{\text{[CLS]}}$ is the classification token representation from the Decoder output and $y$ is the fault type label. The Decoder further generates a causal explanation chain in an autoregressive manner, i.e., a natural language description of the hop-by-hop traceback path from end-point alerts to the root-cause device, with the generation process constrained by the causal structure encoded in $\mathbf{Z}^{(L)}$.

The generated causal chain undergoes consistency verification against the dual-layer KG. For each claim, three conditions are checked: topological connectivity in $\mathcal{G}_{\text{infra}}$, temporal conformity with propagation direction, and historical precedent in $\mathcal{G}_{\text{fault}}$. Failing claims trigger re-generation under stronger knowledge constraints, suppressing hallucination. The root-cause device is localized as the terminal node where traceback finds no further upstream predecessor.

\section{Experiments}

We conduct systematic experiments to answer four research questions:

\textbf{RQ1} evaluates overall performance on both fault type diagnosis and root cause localization, and reveals the causal linkage between them through conditional analysis.

\textbf{RQ2} examines the causal reasoning capability that TCPA confers on the model from the perspective of multi-hop cascading scenario analysis.

\textbf{RQ3} quantifies the independent contribution and synergistic effects of each component through ablation studies, and further investigates the cold-start scenario performance and inference efficiency.

\textbf{RQ4} delves into per-class performance across fault types, combining confusion matrices, misclassification patterns, and a diagnostic case study to reveal the mechanism by which hop-by-hop backtracking resolves end-point ambiguity.

\subsection{Experimental settings}
\subsubsection{Datasets}

The main experiments are conducted on the Wi-Fi Multimodal Fault Benchmark dataset~\cite{zhang2026realisticwififaultdiagnosis}. Built on a real physical testbed with 3 Basic Service Sets and 21 nodes in an AP-STA tree topology, the dataset contains approximately 600 fault cases covering 11 fault types and 1 normal state, with roughly balanced samples per class. Each case provides four types of multimodal monitoring data—traffic-level metrics, packet-level traces, alert events, and system logs—totaling 18,475 alert records. The dataset is temporally split, with the first 80\% used for training and KG construction and the remaining 20\% for testing.

Supplementary experiments are conducted on the TeleLogs dataset~\cite{telelogs}. TeleLogs targets root cause analysis in 5G wireless networks by simulating drive-test scenarios based on real network engineering parameters, with user equipment moving across multiple gNodeB base stations. It includes eight root cause types covering typical 5G faults such as downtilt misconfiguration, co-channel interference, PCI conflict, and handover threshold errors. Each case provides two modalities: base station configuration parameters and user-plane measurement metrics.
Compared with the primary Wi-Fi dataset, TeleLogs features a different fault propagation mechanism from hop-by-hop tree topology to interference and handover propagation between base stations, resulting in substantial differences in both topology structure and fault patterns. Data partitioning follows the official TeleLogs standard.

Adapting PropLLM to TeleLogs involves three scenario-specific configurations: the $G_\text{infra}$ layer replaces the Wi-Fi device dependency graph with a 5G base station neighbor relation graph; the causal direction masks in TCPA replace tree depth with the directional order of inter-base-station interference propagation; and the propagation path annotations in $G_\text{fault}$ replace alarm temporal chains with inter-base-station interference and handover anomaly propagation paths. The core architecture and training configuration remain unchanged.

\subsubsection{Implementation Details}

PropLLM uses Qwen3-8B~\cite{yang2025qwen3} as the backbone language model, with the LLM Decoder initialized from pretrained weights and most parameters frozen, unfreezing only the classification head, cross-attention layers, and LoRA adapters. The TCPA Transformer is configured with 6 layers, 8 attention heads per layer, hidden dimension $d{=}512$, truncation order $K{=}4$, and temporal decay coefficient $\lambda{=}2.0$. The LSTM encoder is a 2-layer bidirectional structure, and the GAT encoder has 2 layers with 4 attention heads. Subgraph retrieval uses neighborhood expansion hops $L{=}2$ and similarity threshold $\tau{=}0.7$.

Training proceeds in two stages. The first stage is supervised fine-tuning with cross-entropy loss for 10 epochs at learning rate 2e-5 with cosine annealing. The second stage is RL fine-tuning with PPO for 5 epochs, reward weights $\alpha{=}1.0$, $\beta{=}0.5$, $\gamma{=}0.3$, clip ratio $\epsilon{=}0.2$, and learning rate 5e-6. All experiments are conducted on 4 NVIDIA A100-80G GPUs with DeepSpeed ZeRO-2 acceleration.

\subsubsection{Baselines}

Baselines for the fault type diagnosis experiment are divided into non-LLM and LLM methods. Non-LLM methods include XGBoost~\cite{chen2016xgboost}, GDN~\cite{deng2021gdn}, Minder~\cite{deng2025minder}, and FAMOS~\cite{duan2025famos}, spanning the technical spectrum from traditional machine learning to deep multimodal fusion. LLM methods include LLM-Direct, LLM-RAG, LLM-CoT, NetLLM~\cite{wu2024netllm}, Confucius~\cite{wang2025confucius}, and BiAn~\cite{wang2025llm}, covering the full paradigm range from zero-shot to specialized architectures. Baselines for the root cause localization experiment include non-LLM methods GDN~\cite{deng2021gdn}, MULAN~\cite{zheng2024mulan}, Chain-of-Event~\cite{yao2024chain}, CORAL~\cite{coral}, RUN~\cite{run}, BARO~\cite{baro}, and AERCA~\cite{aerca}, as well as LLM methods LLM-RAG, LLM-CoT, and BiAn~\cite{wang2025llm}. To ensure fair comparison, all LLM-based methods uniformly adopt Qwen3-8B as the backbone language model; for methods originally designed with different backbone models, we re-implement them with Qwen3-8B while keeping all other configurations consistent with their original papers.

\subsection{RQ1: Overall Performance}
Tables~\ref{tab:diagnosis} and~\ref{tab:rca} report the overall results for fault type diagnosis and root cause localization, respectively. PropLLM achieves 91.2\% Acc and 74.8\% RC@1, outperforming all baselines by a clear margin---3.9\% and 4.7\% above the strongest baseline BiAn, respectively. 

\begin{table}[h]
  \centering
  \caption{Overall fault type diagnosis performance (mean $\pm$ std over 3 independent runs). Precision, Recall, and F1
  are macro-averaged.}
  \label{tab:diagnosis}
  \resizebox{\columnwidth}{!}{%
  \begin{tabular}{llcccc}
  \toprule
  \textbf{Category} & \textbf{Method} & \textbf{Acc(\%)} & \textbf{Precision} & \textbf{Recall} & \textbf{F1} \\
  \midrule
  \multirow{4}{*}{Non-LLM}
  & XGBoost        & 65.3              & 0.642              & 0.618           & 0.630 \\
  & GDN            & 71.2$\pm$0.6      & 0.700$\pm$.005     & 0.678$\pm$.006  & 0.689$\pm$.005 \\
  & Minder         & 83.9$\pm$0.5      & 0.830$\pm$.006     & 0.814$\pm$.007  & 0.822$\pm$.006 \\
  & FAMOS          & 86.1$\pm$0.5      & 0.851$\pm$.006     & 0.839$\pm$.007  & 0.845$\pm$.006 \\
  \midrule
  \multirow{6}{*}{LLM}
  & LLM-Direct     & 57.6$\pm$1.2      & 0.556$\pm$.015     & 0.519$\pm$.016  & 0.537$\pm$.014 \\
  & LLM-RAG        & 74.6$\pm$0.8      & 0.740$\pm$.009     & 0.714$\pm$.010  & 0.727$\pm$.009 \\
  & LLM-CoT        & 78.0$\pm$0.9      & 0.771$\pm$.010     & 0.749$\pm$.011  & 0.760$\pm$.010 \\
  & NetLLM         & 84.7$\pm$0.7      & 0.839$\pm$.008     & 0.823$\pm$.009  & 0.831$\pm$.008 \\
  & Confucius      & 85.3$\pm$0.6      & 0.844$\pm$.007     & 0.830$\pm$.008  & 0.837$\pm$.007 \\
  & BiAn           & 87.3$\pm$0.6      & 0.865$\pm$.007     & 0.851$\pm$.008  & 0.858$\pm$.007 \\
  \midrule
  & \textbf{PropLLM (Ours)} & \textbf{91.2$\pm$0.4} & \textbf{0.910$\pm$.005} & \textbf{0.896$\pm$.006} &
  \textbf{0.903$\pm$.005} \\
  \bottomrule
  \end{tabular}%
  }
\end{table}

\begin{table}[t]
  \centering
  \caption{Root cause localization performance (mean $\pm$ std, 3 independent runs).}
  \label{tab:rca}
  \renewcommand{\arraystretch}{1.1}
  \resizebox{\columnwidth}{!}{%
  \begin{tabular}{l|cc|cc}
  \toprule
  \multirow{2}{*}{Method} & \multicolumn{2}{c|}{Wi-Fi Multimodal Fault Dataset} & \multicolumn{2}{c}{TeleLogs (5G)} \\
  \cmidrule{2-5}
   & RC@1 (\%) & RC@3 (\%) & RC@1 (\%) & RC@3 (\%) \\
  \midrule
  GDN & 46.2$\pm$0.9 & 64.7$\pm$0.8 & 71.5$\pm$0.9 & 84.3$\pm$0.8 \\
  MULAN & 53.0$\pm$0.8 & 71.5$\pm$0.7 & 76.8$\pm$0.8 & 88.5$\pm$0.7 \\
  Chain-of-Event & 56.5$\pm$0.8 & 73.8$\pm$0.7 & 79.6$\pm$0.8 & 90.3$\pm$0.7 \\
  CORAL & 57.3$\pm$0.7 & 74.8$\pm$0.6 & 80.2$\pm$0.7 & 90.8$\pm$0.6 \\
  RUN & 60.2$\pm$0.7 & 77.4$\pm$0.6 & 83.4$\pm$0.7 & 93.0$\pm$0.6 \\
  BARO & 62.8$\pm$0.6 & 79.5$\pm$0.5 & 85.7$\pm$0.6 & 94.5$\pm$0.5 \\
  AERCA & 65.1$\pm$0.6 & 81.0$\pm$0.5 & 88.2$\pm$0.5 & 95.8$\pm$0.5 \\
  \midrule
  LLM-RAG & 50.5$\pm$1.0 & 69.0$\pm$0.9 & 74.6$\pm$0.9 & 87.0$\pm$0.8 \\
  LLM-CoT & 58.9$\pm$0.9 & 76.3$\pm$0.8 & 82.5$\pm$0.8 & 92.4$\pm$0.7 \\
  BiAn & 70.1$\pm$0.7 & 85.4$\pm$0.6 & 92.8$\pm$0.4 & 97.8$\pm$0.3 \\
  \midrule
  PropLLM (Ours) & \textbf{74.8$\pm$0.5} & \textbf{89.6$\pm$0.4} & \textbf{96.3$\pm$0.3} & \textbf{99.1$\pm$0.2} \\
  \bottomrule
  \end{tabular}%
  }
\end{table}

The performance hierarchy in Table~\ref{tab:diagnosis} reveals three noteworthy observations:

\emph{First}, general-purpose LLM parametric knowledge is insufficient for network fault diagnosis: LLM-Direct achieves only 57.6\%, below XGBoost's 65.3\%, confirming that cascading fault propagation exceeds zero-shot capability. From LLM-RAG through LLM-CoT to NetLLM, external knowledge, chain-of-thought guidance, and domain adaptation contribute stackable gains, yet all operate on the input or decoding side without modifying the model's internal perception of causal direction.

\emph{Second}, the gap between non-LLM and LLM methods is not a simple generational divide. FAMOS achieves 86.1\% Acc, surpassing NetLLM and Confucius through effective cross-modal fusion, but its one-shot mapping paradigm cannot reason stepwise along propagation paths. BiAn breaks through the non-LLM ceiling at 87.3\% through hierarchical summarization and multi-step reasoning.

\emph{Third}, PropLLM's 3.9\% Acc gain over BiAn quantifies the contribution of causally-aware attention and hop-by-hop fact verification. BiAn's summarization--reasoning pipeline lacks topological causal direction constraints and per-hop fact checking. PropLLM's TCPA injects propagation direction priors at the attention level and cross-validates with the dual-layer KG at every hop, producing a synchronized 4.5\% improvement in both Precision
(filtering spurious paths) and Recall (discovering discriminative evidence at intermediate nodes).

In the root cause localization dimension, PropLLM achieves 74.8\% RC@1 on the Wi-Fi Multimodal Fault Dataset, improving over BiAn by 4.7\%, and 96.3\% RC@1 on TeleLogs, improving over BiAn by 3.5\%. The RC@1 gains exceed those in accuracy on both datasets. This is not coincidental: root cause localization demands precise backtracking along the full propagation chain, where even small causal direction errors are highly detrimental. Thus, the constraints imposed by TCPA’s causal masks are particularly effective.
All methods perform substantially better on TeleLogs than on the Wi-Fi dataset, consistent with TeleLogs’ clearer causal signals from well-defined configuration deviations for each of its eight root causes versus the intermediate state aliasing caused by multi-hop cascading in Wi-Fi. The notably high 96.3\% RC@1 on TeleLogs further demonstrates PropLLM’s cross-topology adaptability, as TCPA effectively captures the dominant propagation direction even when switching from tree structures to 5G inter-base-station interference patterns.
Among non-LLM methods, Chain-of-Event reaches 56.5\% RC@1 on Wi-Fi by learning weighted event causal graphs, outperforming MULAN but falling slightly below CORAL. AERCA further improves to 65.1\% via Granger causal discovery. However, these approaches rely on learning causal structures from limited observational data. In contrast, PropLLM directly encodes known causal priors from topology in $  \mathcal{G}_{\text{infra}}  $ and historical fault templates in $  \mathcal{G}_{\text{fault}}  $, bypassing the statistical bottleneck of data-driven causal graph learning.

\begin{figure}[h]
  \centering
  \includegraphics[width=0.8\columnwidth]{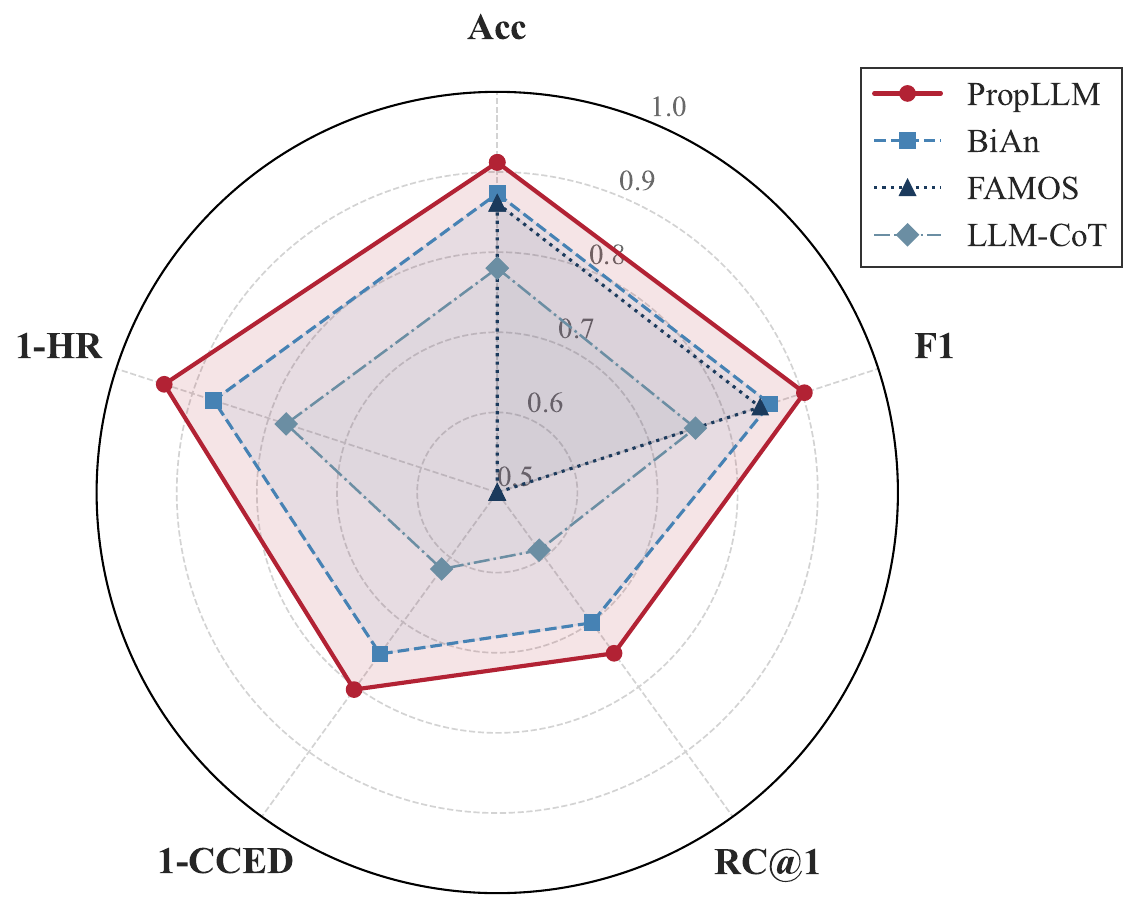}
  \caption{Multi-dimensional performance radar chart of PropLLM and representative baselines.}
  \label{F4.pdf}
  \end{figure}

Fig.~\ref{F4.pdf} presents a radar chart across five dimensions: Acc, F1, RC@1, 1-CCED, and 1-HR (the latter two mapped to $[0,1]$ via min-max normalization). PropLLM occupies the outermost contour on all axes, with the largest lead over BiAn on the 1-HR and RC@1 axes---directly reflecting that the core benefit of hop-by-hop verification lies in causal chain fidelity and root cause precision rather than classification accuracy alone.

\begin{figure}[h]
  \centering
  \includegraphics[width=0.8\columnwidth]{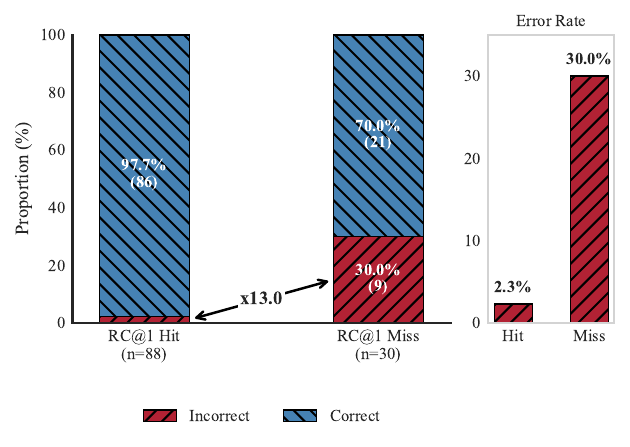}
  \caption{Conditional association between root cause localization correctness and classification accuracy.}
  \label{F5.pdf}
  \end{figure}

\textbf{Causal propagation analysis between root cause localization and classification.} The results above demonstrate that PropLLM significantly outperforms baselines on both dimensions, but a deeper question remains: does root cause localization accuracy truly drive classification performance? To validate this core thesis, we partition PropLLM's test cases on the Wi-Fi Multimodal Fault Dataset (118 cases) into two groups according to whether RC@1 hits, and compute the classification accuracy for each group. The results are shown in Fig.~\ref{F5.pdf}.

The RC@1-hit group achieves a classification accuracy of 97.7\%, while the miss group drops sharply to 70.0\%, with the error rate surging from 2.3\% to 30.0\%. The only 2 classification errors in the hit group both originate from complex topologies where multiple propagation paths converge---even though the root cause device is correctly located, the mixed symptoms downstream of the convergence point still cause type confusion, representing the most challenging cases in this dataset. The 30.0\% error rate in the miss group reveals a causal amplification effect: a root cause localization error is not an isolated local mistake but rather corrupts the starting point of every subsequent hop-by-hop fact verification, causing deviations to accumulate along the causal chain until classification judgment loses its reliable basis. This result provides empirical evidence that the synchronized improvement of PropLLM on Acc and RC@1 is not a coincidence across two independent dimensions, but rather the inevitable consequence of root cause localization accuracy propagating through the causal chain to the classification decision.

\subsection{RQ2: Multi-hop cascading scenario analysis}

We partition the fault cases in the test set into three groups by causal chain length after excluding the Normal class, yielding 108 cases total: short chains with 1--2 hops containing 46 cases, medium chains with 3--4 hops containing 42 cases, and long chains with 5 or more hops containing 20 cases. Table~\ref{tab:multihop} compares PropLLM against representative baselines that possess both root cause localization capability and causal reasoning mechanisms.

\begin{table}[h]
  \centering
  \caption{Acc (\%) and RC@1 (\%) across different propagation hop counts (mean $\pm$ std, 3 independent runs).}
  \label{tab:multihop}
  \resizebox{\columnwidth}{!}{
  \begin{tabular}{lcccccc}
  \toprule
  \multirow{2}{*}{Method} & \multicolumn{2}{c}{Short (1--2 hops, $n$=46)} & \multicolumn{2}{c}{Medium (3--4 hops,
  $n$=42)} & \multicolumn{2}{c}{Long ($\geq$5 hops, $n$=20)} \\
  \cmidrule(lr){2-3} \cmidrule(lr){4-5} \cmidrule(lr){6-7}
   & Acc & RC@1 & Acc & RC@1 & Acc & RC@1 \\
  \midrule
  MULAN & 87.0$\pm$1.4 & 63.0$\pm$1.6 & 77.0$\pm$1.5 & 50.0$\pm$1.8 & 60.0$\pm$2.5 & 34.0$\pm$2.8 \\
  AERCA & 90.0$\pm$1.1 & 75.0$\pm$1.3 & 82.0$\pm$1.3 & 64.0$\pm$1.5 & 66.0$\pm$2.2 & 46.5$\pm$2.5 \\
  BARO & 89.0$\pm$1.2 & 71.0$\pm$1.4 & 81.0$\pm$1.4 & 59.5$\pm$1.6 & 65.0$\pm$2.3 & 42.5$\pm$2.7 \\
  BiAn & 92.5$\pm$0.9 & 79.5$\pm$1.1 & 86.5$\pm$1.1 & 69.0$\pm$1.3 & 72.0$\pm$2.0 & 51.5$\pm$2.3 \\
  PropLLM (Ours) & \textbf{94.0$\pm$0.7} & \textbf{83.0$\pm$0.9} & \textbf{90.5$\pm$0.8} & \textbf{74.0$\pm$1.0} &
  \textbf{82.0$\pm$1.5} & \textbf{57.5$\pm$1.8} \\
  \bottomrule
  \end{tabular}
  }
\end{table}

All methods degrade as hop count increases, but their decay rates differ fundamentally. PropLLM's Acc decreases by 12.0\% from short to long chains, compared with BiAn's 20.5\% decrease, yielding a decay rate only 59\% of the latter. In the RC@1 dimension, the gap between PropLLM and BiAn widens from 3.5\% on short chains to 6.0\% on long chains, with the advantage growing approximately linearly with chain length. The underlying logic of this trend is that short-chain scenarios present limited path ambiguity where one-shot reasoning already handles effectively and the marginal gain of hop-by-hop verification is modest. As chain length increases, each hop introduces new branching possibilities, causing methods without causal direction constraints to rapidly degrade under cumulative ambiguity. The TCPA causal masks eliminate reverse attention at each hop, keeping ambiguity accumulation within a manageable range.

\begin{figure}[h]
  \centering
  \includegraphics[width=1\columnwidth]{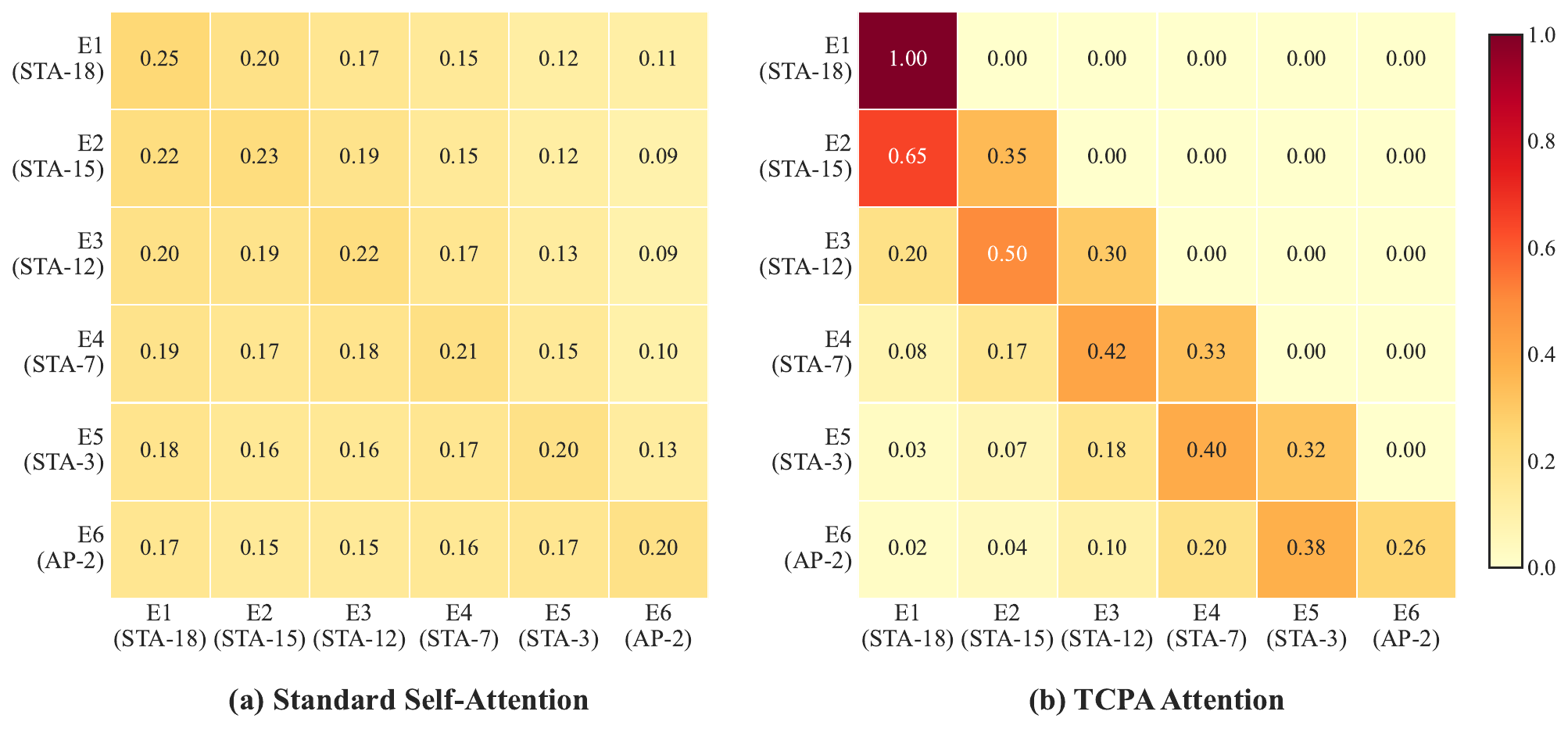}
  \caption{Attention weight matrices of standard self-attention (left) vs.\ TCPA (right) on a 5-hop long-chain case.}
  \label{F6.pdf}
\end{figure}

Fig.~\ref{F6.pdf} reveals the mechanistic source through attention heatmaps of a 5-hop case. The standard self-attention matrix is approximately symmetric, with the end-point STA's throughput alarm receiving the highest weight, causing misidentification of this symptom as the root cause. The TCPA matrix exhibits a clear lower-triangular structure---the causal mask zeroes out all reverse attention, forcing the model to look exclusively upstream. A gradient band along the diagonal reflects the diffusion matrix $\exp(-\beta \mathbf{L})$: weights decay exponentially with topological distance, concentrating focus on causal neighbors. These two features explain why PropLLM decays most slowly in the long-chain regime in Table~\ref{tab:multihop}: even beyond 5 hops, attention at each hop remains constrained to the correct causal direction.

\subsection{RQ3: Ablation Study}

Table~\ref{tab:ablation} quantifies the independent contribution of each component by removing them one at a time.

\begin{table}[h]
  \centering
  \caption{Ablation study results.}
  \label{tab:ablation}
  \resizebox{\columnwidth}{!}{%
  \begin{tabular}{lcccc}
  \toprule
  \textbf{Variant} & \textbf{Acc(\%)} & \textbf{F1} & \textbf{RC@1(\%)} & \textbf{HR(\%) ($\downarrow$)} \\
  \midrule
  PropLLM (Full)                        & 91.2 & 0.903 & 74.8 & 6.3  \\
  w/o TCPA (standard self-attention)    & 81.9 & 0.799 & 58.7 & 22.7 \\
  w/o causal direction mask             & 86.2 & 0.846 & 66.4 & 14.1 \\
  w/o propagation diffusion matrix      & 87.0 & 0.857 & 67.2 & 12.8 \\
  w/o temporal propagation bias         & 88.7 & 0.874 & 70.6 & 9.5  \\
  w/o dual-layer KG (single-layer KG)   & 85.3 & 0.834 & 62.9 & 15.6 \\
  w/o $\mathcal{G}_\text{fault}$ (only $\mathcal{G}_\text{infra}$) & 87.6 & 0.862 & 66.9 & 13.2 \\
  w/o cross-layer links                 & 86.5 & 0.850 & 65.1 & 14.5 \\
  w/o hop-by-hop retrieval (one-shot RAG) & 84.4 & 0.823 & 61.3 & 18.3 \\
  w/o RL fine-tuning (SFT only)         & 87.9 & 0.864 & 69.7 & 11.8 \\
  w/o multi-modal (text-only input)     & 83.6 & 0.816 & 60.4 & 9.1  \\
  \bottomrule
  \end{tabular}%
  }
  \end{table}

The ablation results can be interpreted along three dimensions. Regarding independent component contributions, TCPA has the most pronounced impact: removing it entirely causes Acc to drop by 9.3\% to 81.9\%, RC@1 to drop by 16.1\% to 58.7\%, and HR to surge from 6.3\% to 22.7\%, with all three dimensions degrading simultaneously. This confirms that causal direction awareness is the cornerstone of the hop-by-hop backtracking paradigm. The three sub-mechanisms within TCPA contribute in a hierarchical fashion. The causal direction mask causes a 5.0\% Acc decline, the propagation diffusion matrix causes a 4.2\% decline, and the temporal bias causes a 2.5\% decline. This pattern is consistent with the design rationale, where direction judgment is the fundamental prerequisite, distance decay refines precision, and temporal bias provides auxiliary correction. On the knowledge side, degrading hop-by-hop retrieval to one-shot RAG causes HR to surge to 18.3\%, the largest HR increase among all ablation variants, indicating that dynamic per-hop factual constraints contribute even more to reasoning faithfulness than RL fine-tuning.

Regarding inter-component synergy, removing TCPA drops Acc to 81.9\%, which falls not only far below the full model but even below FAMOS at 86.1\% and BiAn at 87.3\%; removing hop-by-hop retrieval and removing multi-modal inputs similarly produce Acc values of 84.4\% and 83.6\% that fall below baseline levels. This reveals that PropLLM's components are not independently pluggable modules but rather form a tightly coupled synergistic system. Without TCPA's directional guidance, hop-by-hop retrieval may backtrack in the wrong direction and inject irrelevant subgraph information, performing worse than one-shot methods. This ``negative transfer'' phenomenon provides a counter-proof of  PropLLM's core thesis: propagation-aware reasoning and hop-by-hop fact verification must be deeply coupled, and the absence of either side substantially undermines the utility of the other.

Additionally, removing multi-modal inputs causes HR to drop to 9.1\%, which is paradoxically lower than 11.8\% when removing RL and 13.2\% when removing $\mathcal{G}_\text{fault}$---a non-monotonic phenomenon worth noting. With only text alarm inputs, the model's reasoning space is constrained to the factual scope of textual descriptions, producing shorter causal chains with an average length dropping from 3.8 to 2.9 hops, which reduces hallucination opportunities but at the cost of a 7.6\% Acc decline and a 14.4\% RC@1 decline, as the model misses critical causal evidence embedded in the temporal and topological modalities while avoiding hallucinations. This indicates that HR should not be interpreted in isolation---high-fidelity reasoning must be built on a sufficient information foundation.

\textbf{Cold-start scenario analysis.}
$\mathcal{G}_\text{fault}$ relies on historical fault cases, so its performance when cases are scarce during early deployment of a new network deserves attention. Table~\ref{tab:coldstart} quantifies PropLLM's performance under varying levels of historical accumulation by controlling the case volume in $\mathcal{G}_\text{fault}$.

\begin{table}[t]
  \centering
  \caption{Impact of $\mathcal{G}_\text{fault}$ scale on performance (cold-start analysis).}
  \label{tab:coldstart}
  \begin{tabular}{lcccccc}
  \toprule
  \textbf{$\mathcal{G}_\text{fault}$ cases} & \textbf{0} & \textbf{10\%} & \textbf{30\%} & \textbf{50\%} & \textbf{80\%}
   & \textbf{100\%} \\
  \midrule
  Acc(\%)  & 87.6 & 88.4 & 89.5 & 90.3 & 91.0 & 91.2 \\
  RC@1(\%) & 66.9 & 68.8 & 71.4 & 73.1 & 74.5 & 74.8 \\
  \bottomrule
  \end{tabular}
  \end{table}

When $\mathcal{G}_\text{fault}$ is completely empty, PropLLM still achieves 87.6\% Acc, on par with BiAn's 87.3\%, indicating that TCPA and $\mathcal{G}_\text{infra}$ alone already match the strongest baseline, and $\mathcal{G}_\text{fault}$ serves as the critical increment for surpassing it rather than a necessary prerequisite. Performance gains exhibit diminishing marginal returns as cases accumulate: just 10\% of historical cases push Acc to 88.4\%, at 30\% it reaches 89.5\% which already approximates 98\% of full performance, and at 80\% it essentially saturates. This rapid convergence stems from the reusability of causal templates---faults of the same type exhibit highly similar propagation patterns even when occurring on different devices, so a small number of cases suffices to cover the principal modes. Furthermore, verified propagation paths from each successful diagnosis can be automatically written back to $\mathcal{G}_\text{fault}$, enabling system performance to improve continuously over operational time. It should be noted that this experiment only controls the case volume in $\mathcal{G}_\text{fault}$ while keeping the SFT training data at full scale; the 87.6\% Acc with an empty $\mathcal{G}_\text{fault}$ can be regarded as an optimistic upper bound on cold-start performance, and a joint cold-start analysis that simultaneously reduces both training data and $\mathcal{G}_\text{fault}$ is left for future work.

\textbf{Inference efficiency analysis.} Table~\ref{tab:latency} reports the average inference latency per diagnostic case for PropLLM and representative baselines (4$\times$ A100-80G, batch size = 1).

\begin{table}[t]
  \centering
  \caption{Average inference latency per diagnostic case.}
  \label{tab:latency}
  \begin{tabular}{lcc}
  \toprule
  Method & Latency (s) & Dominant Component \\
  \midrule
  FAMOS & 0.08$\pm$0.01 & Single forward pass \\
  LLM-Direct & 0.35$\pm$0.05 & Single LLM generation \\
  Chain-of-Event & 0.42$\pm$0.06 & Causal graph + inference \\
  LLM-CoT & 0.82$\pm$0.12 & Single LLM generation (w/ CoT) \\
  PropLLM (Ours) & 1.23$\pm$0.15 & Hop-by-hop retrieval + TCPA + LLM \\
  BiAn & 1.45$\pm$0.18 & Two-stage LLM generation \\
  \bottomrule
  \end{tabular}
\end{table}

PropLLM achieves an average inference latency of 1.23 seconds, lower than BiAn's 1.45 seconds. Latency scales approximately linearly with chain length: short-chain (1--2 hops) 0.72s, medium-chain (3--4 hops) 1.35s, and long-chain (5+ hops) 1.84s. The LLM Decoder's autoregressive generation dominates at approximately 75\% of total latency, while TCPA and KG retrieval together account for only about 10\%, indicating negligible overhead from causality-aware attention. TCPA's complexity is $O(N^2 L)$, which is negligible at the current scale ($N{=}21$), and the diffusion matrix $D$ is precomputed offline. For larger networks, the $L$-hop neighborhood expansion confines retrieval to local topology, making latency weakly dependent on total network scale. Compared with manual troubleshooting (minutes to hours), PropLLM's second-level latency fully satisfies online assisted diagnosis requirements.

\subsection{RQ4: Fine-Grained Analysis and Case Study}

\textbf{Per-class diagnosis performance.}
Table~\ref{tab:perclass} reports the per-class F1 scores of PropLLM across all 12 categories compared with the strongest baseline BiAn.

\begin{table}[t]
  \centering
  \caption{Per-class F1 scores across 12 fault types.}
  \label{tab:perclass}
  \begin{tabular}{lccc}
  \toprule
  \textbf{Fault Type} & \textbf{PropLLM} & \textbf{BiAn} & \textbf{$\Delta$} \\
  \midrule
  AppCrash        & 0.941 & 0.920 & +0.021 \\
  AppSlowdown     & 0.842 & 0.765 & +0.077 \\
  BeaconLoss      & 0.952 & 0.930 & +0.022 \\
  BufferBloat     & 0.857 & 0.835 & +0.022 \\
  HiddenNode      & 0.824 & 0.722 & +0.102 \\
  NodeCrash       & 0.941 & 0.897 & +0.044 \\
  Normal          & 1.000 & 0.965 & +0.035 \\
  PoorLinkQuality & 0.842 & 0.765 & +0.077 \\
  ProbeFailure    & 0.923 & 0.878 & +0.045 \\
  QueueOverflow   & 0.875 & 0.847 & +0.028 \\
  RateAdaptation  & 0.900 & 0.848 & +0.052 \\
  TrafficOverload & 0.941 & 0.924 & +0.017 \\
  \bottomrule
  \end{tabular}
\end{table}

PropLLM outperforms BiAn across all 12 categories. However, the magnitude of improvement is highly uneven and closely aligns with the method’s design expectations. The largest gains occur in HiddenNode, AppSlowdown, and PoorLinkQuality (F1 improvements of 0.102, 0.077, and 0.077, respectively). These categories have highly confusable terminal symptoms—all manifesting as decreased throughput and increased latency at end nodes—making them nearly indistinguishable from terminal alerts alone. PropLLM’s hop-by-hop traceback effectively resolves this by mining discriminative state features at intermediate nodes.
By contrast, the smallest improvements are observed in TrafficOverload, AppCrash, and BeaconLoss (gains of only 0.017 to 0.022). Their terminal symptoms already possess extremely high discriminability, so the existing one-shot mapping approach is sufficient and leaves limited marginal benefit for hop-by-hop traceback.

\begin{figure}[t]
  \centering
  \includegraphics[width=\columnwidth]{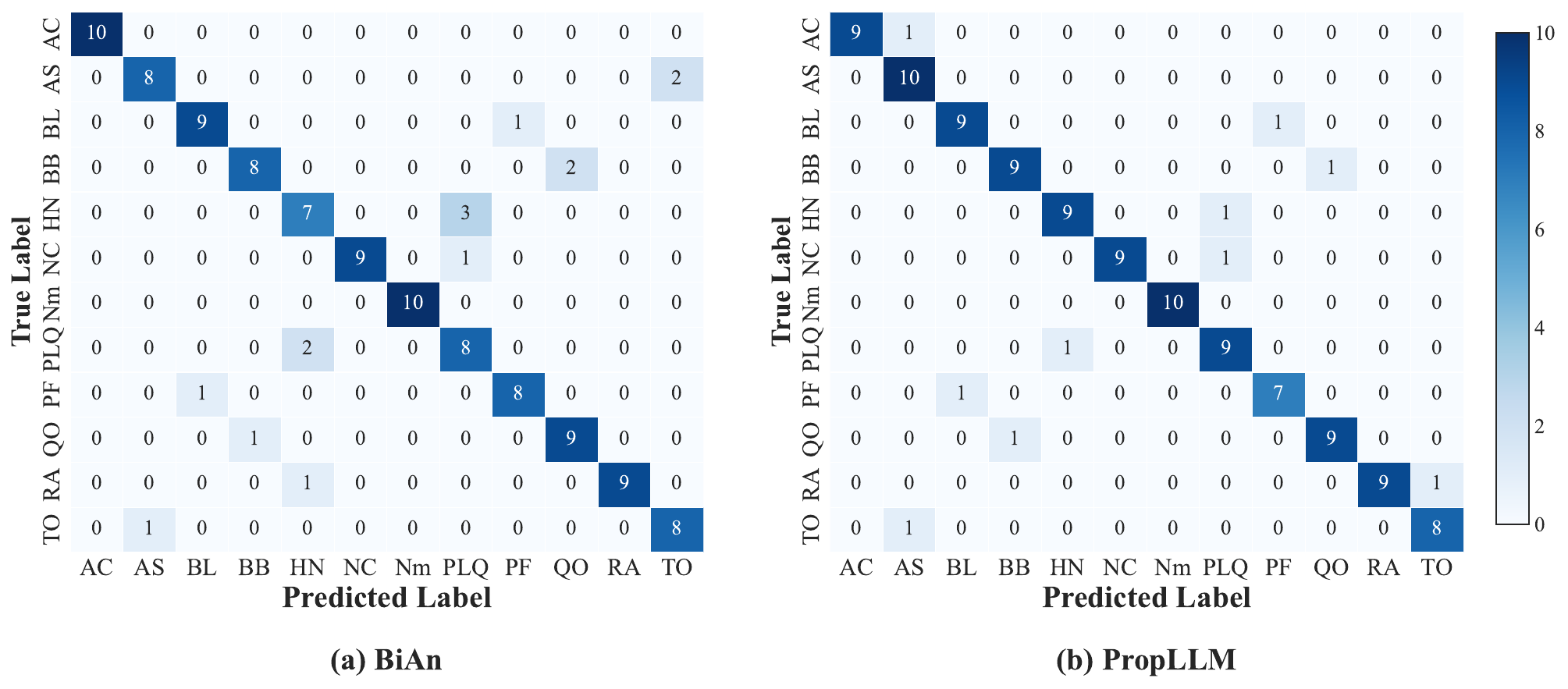}
  \caption{Confusion matrix heatmaps of BiAn (left) and PropLLM (right).}
  \label{F8.pdf}
\end{figure}

Fig.~\ref{F8.pdf} displays the 12-class confusion matrices of BiAn and PropLLM side by side as heatmaps. BiAn’s confusion matrix exhibits prominent off-diagonal blocks in three category-pair regions: HiddenNode/PoorLinkQuality, AppSlowdown/TrafficOverload, and BufferBloat/QueueOverflow. These blocks show concentrated and symmetric misclassifications that reflect the inherently bidirectional ambiguity of end-point symptoms. PropLLM’s confusion matrix is highly diagonalized, with the off-diagonal elements in these three regions nearly zeroed out. The residual 10 misclassifications are sparsely distributed with no discernible clustering pattern, indicating a qualitative shift from “systematic category confusion” to “case-level sporadic errors.”

\begin{figure}[h]
  \centering
  \includegraphics[width=\columnwidth]{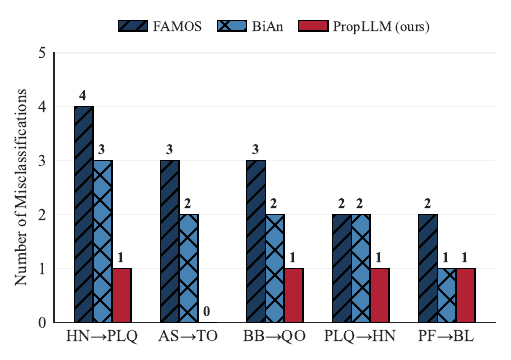}
  \caption{Misclassification counts on top-5 confusion pairs.}
  \label{F9.pdf}
\end{figure}

\textbf{Misclassification pattern analysis.}
Fig.~\ref{F9.pdf} compares the misclassification counts of PropLLM, BiAn, and FAMOS on the top-5 confusion pairs using grouped bar charts.

Baseline misclassifications concentrate heavily on category pairs with similar end-point symptoms: the top four confusion pairs account for 9 out of BiAn's total 15 errors, i.e., 60\%, and FAMOS concentrates 12 out of 16, i.e., 75\%. PropLLM produces only 4 misclassifications across these five confusion pairs, less than half of BiAn's count, because hop-by-hop backtracking acquires discriminative evidence at intermediate nodes that end-point alarms cannot provide. PropLLM's residual 10 errors are uniformly dispersed across category pairs with no more than one per pair, occurring primarily in atypical propagation patterns that are rare in the training set. As $\mathcal{G}_\text{fault}$ cases continue to accumulate, these sporadic misclassifications will progressively diminish.

\begin{figure}[h]
  \centering
  \includegraphics[width=\columnwidth]{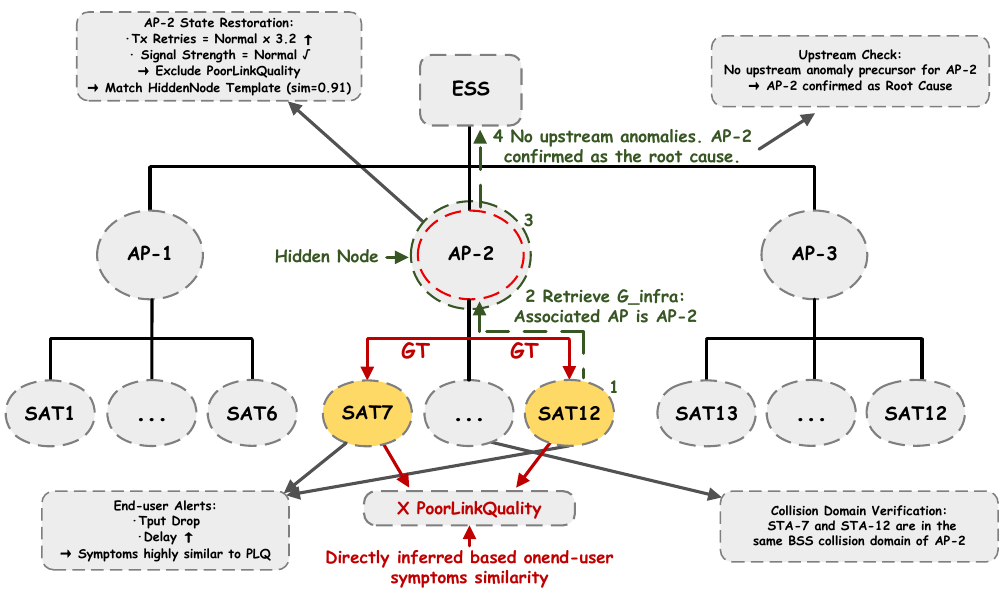}
  \caption{Hop-by-hop backtracking path comparison on a HiddenNode fault case. BiAn misclassifies based on end-point
  symptom similarity, while PropLLM correctly backtracks through intermediate nodes to identify the root cause.}
  \label{F10.pdf}
\end{figure}

\textbf{Diagnostic case study.}
Fig.~\ref{F10.pdf} visualizes the hop-by-hop backtracking process on a HiddenNode fault test case, showing the ESS network topology with three APs and associated STAs. Different colored directed arrows mark the ground-truth propagation path (gray dashed), PropLLM's backtracking path (green solid), and BiAn's judgment path (red solid).

In this case, STA-7 and STA-12 simultaneously report throughput degradation and latency increase; BiAn classifies this as PoorLinkQuality based on end-point symptom similarity. PropLLM's trajectory differs: at hop~1, it departs from STA-12 and retrieves from $\mathcal{G}_\text{infra}$ that its associated AP is AP-2; at hop~2, it finds AP-2's retransmission rate is 3.2$\times$ normal while RSSI remains normal---key evidence distinguishing HiddenNode (MAC-layer collisions) from PoorLinkQuality (physical-layer degradation), invisible at end-points; at hop~3, it matches ``high retransmission + normal RSSI'' against $\mathcal{G}_\text{fault}$ causal templates (similarity 0.91) and confirms via $\mathcal{G}_\text{infra}$ that STA-7 and STA-12 share AP-2's collision domain; at hop~4, it verifies no upstream anomalies above AP-2, localizing it as the root cause. This exemplifies the synergy of three components: TCPA ensures upstream backtracking, $\mathcal{G}_\text{infra}$ provides protocol-level facts, and $\mathcal{G}_\text{fault}$ supplies experiential evidence.

\section{Conclusion}
This paper proposes PropLLM, the first framework integrating hop-by-hop scene reconstruction with LLMs for network fault diagnosis via causal backtracking. Its core consists of three tightly coupled components: a dual-layer knowledge graph separating structural and experiential knowledge for verifiable per-hop reconstruction; TCPA, which encodes topological causal priors into attention via causal masks, propagation diffusion matrices, and temporal biases; and a retrieval-verification loop coupling backtracking with dynamic retrieval to suppress hallucinations. Experiments show PropLLM significantly outperforms all baselines on the Wi-Fi Multimodal Fault Dataset and TeleLogs 5G dataset, with pronounced gains in multi-hop scenarios. The main limitation is that experiments cover only IoT wireless and 5G cellular networks; effectiveness on more complex topologies such as data center fat-trees and WAN meshes remains to be verified.

As for future work, we will explore extension to larger-scale heterogeneous network topologies, few-shot cold-start strategies, and model distillation with inference acceleration.

\bibliographystyle{IEEEtran}
\bibliography{ref}


\end{document}